\documentclass[letterpaper]{article} 
\usepackage{aaai2026}  
\usepackage{times}  
\usepackage{helvet}  
\usepackage{courier}  
\usepackage[hyphens]{url}  
\usepackage{graphicx} 
\usepackage{natbib}  
\usepackage{caption} 
\usepackage{algorithm}
\usepackage{algorithmic}
\usepackage{amssymb}
\usepackage{amsmath}
\usepackage{subcaption}

\usepackage{newfloat}
\usepackage{listings}
\DeclareCaptionStyle{ruled}{labelfont=normalfont,labelsep=colon,strut=off} 
\floatstyle{ruled}
\newfloat{listing}{tb}{lst}{}
\floatname{listing}{Listing}
\title{Syntactic Segmentation and Collage Generation for Videogame Levels with Perceptual Distance}
\author{
    Ross Mawhorter,
    Matthew Guzdial
}
\affiliations{
    University of Alberta\\
    116 Street and 85th Avenue\\
    Edmonton, AB\\
    T6G 2R3\\
    mawhorte@ualberta.ca
}

\begin{document}

\maketitle

\begin{abstract}
In procedural videogame level design, many existing techniques combine hand-authored level chunks. Some existing analysis techniques also rely on a higher-level graph structure which abstracts the level design. In this paper, we explore techniques for computing both of these things automatically from an existing design. We introduce the concept of \emph{perceptual distance}, based on analysis techniques from the architecture subfield of Space Syntax. We use this perceptual distance to segment level designs with hierarchical clustering. We compare with past work on \emph{StarCraft} analysis, show results on rooms from \emph{Super Metroid}, and also demonstrate a proof-of-concept level generation algorithm based on this segmentation method.
\end{abstract}


\section{Introduction}

Segmentation is a form of analysis that involves both splitting up something (partitioning), and determining the relationships between the partitions. The quality of a given segmentation is a function of \emph{why} that segmentation was created in the first place. For example, prior work on \emph{StarCraft}~\cite{starcraft} treats tactical analysis as a segmentation problem~\cite{perkins2010terrain}. In this case, clusters denote the partitions of a \emph{StarCraft} map, and the edges between clusters denote choke points in the map: places that a player can use to control the flow of units. The quality of a segmentation for this task depends on how useful it is to the AI systems that depend on it for unit movement.
This paper presents a novel method for 2D level design segmentation based on a concept called \emph{perceptual flux}, which is closely related to isovist fields~\cite{benediktisovist}.

 An isovist is a polygon that describes the visible space at a given location. Perceptual flux measures changes in the boundary of this shape, and our segmentation approach marks segment boundaries based on those changes. This technique is an automated way to address some of the the hand-authoring costs of other procedural level design techniques. We identify three major categories of procedural level design to which our approach could apply. 
Firstly, many existing procedural level design approaches attempt to replicate input data (for example, Markov Chains~\cite{summerville2018procedural} and WaveFunction Collapse (WFC)~\cite{wfc}). These approaches successfully replicate short-range relationships between tiles but often fail to capture longer-range dependencies. 
Existing attempts to solve this problem involve hand-authoring: either adding additional constraints to a constraint-solving implementation of WFC or manually creating a separate set of per-structure rules~\cite{turtle-loves-pizza, nutwfc, hierarchicalwfc}.
A high quality automated segmentation approach could solve such problems without additional human authoring, though we do not apply our approach to this task in this paper.

Other existing procedural level design techniques (for example mission-graph-based generation~\cite{missions} and constrained graph generation from examples~\cite{graphgen}) rely on generating a higher-level graphical description of a level before converting it into actual level geometry. This type of generation relies on either hand-authored grammar rules or hand-authored example graphs. Our segmentation technique is an automated way to create a graph abstraction of an existing tile representation.

Finally, many existing generators work by stitching together pre-authored (or partially pre-authored) content. This creates a different authoring challenge: creating a set of components that can be combined to create many distinct and larger level designs. In this paper we explore this type of collage-based procedural level design using components that are automatically identified by perceptual segmentation.

This paper's contributions are:
\begin{itemize}
    \item A new concept of perceptual distance based on a scaled version of isovist information (along with a justification for why this is more useful than standard isovist statistics).
    \item \emph{Perceptual Distance Segmentation} (PDS): a hierarchical clustering algorithm based on this distance metric (along with comparison to prior work).
    \item A basic collage-based procedural level design algorithm that uses the clustering algorithm to automatically remix existing content (and an example application to \emph{Super Metroid}~\cite{super_metroid}).
\end{itemize}

\section{Background}

This paper draws on prior work applying computational geometry techniques to analyzing spaces (whether those spaces are floor plans of actual buildings or videogame levels). This also extends a line of work that uses architectural analysis techniques on videogames. Boakye et al. use Lynch's theory of cities to describe open world games~\cite{imgopenworld, lynch1964image}. Salge and Herv\'{e} show that some isovist information correlates with human evaluation of procedurally-generated Minecraft settlements~\cite{salgeisovist}.

\subsection{Level Segmentation}

The primary prior work that focuses specifically on videogame level design segmentation is Perkins's work on \emph{StarCraft: Brood War} tactical analysis (BWTA)~\cite{perkins2010terrain}. They use a line-segment Voronoi diagram~\cite{karavelas2004robust} to identify the basic structure of the space, prune it to identify the centers of important regions. 
There are similarities between BWTA and PDS (our algorithm). Both algorithms initially identify regions by selecting important points within them (BWTA uses the relative size of the points; PDS uses the minima of the perceptual flux). They then merge these regions (BWTA uses a custom merging rule; PDS uses hierarchical clustering). Finally, both algorithms identify the boundaries between the merged regions (BWTA uses the smallest diameter while PDS uses the perceptual distance to assign points to a region).

Some of the advantages of BWTA for tactical analysis (e.g. identifying the locations of specific choke points) are disadvantages when it comes to identifying an appropriate high-level structure of a given map (for example, room boundaries are not necessarily straight lines). The existence of a small solid area within a room necessarily creates additional regions in the Delaunay graph (and is important from the perspective of tactical analysis), but does not necessarily have to break apart a room (from the perspective of graphical structure). The presence of this solid island should change the way you arrange your units for battle in \emph{StarCraft}, but does not necessarily create a separate region for analysis when it comes to design tasks.


\subsection{Isovist Analysis and Space Syntax}

Benedikt's seminal paper introducing the concept of isovists (and associated isovist fields) has been the basis for many techniques for analyzing spaces (a subfield called Space Syntax)~\cite{benediktisovist}. The core question of this technique of analysis is how the set of visible points changes as a perceiver moves through a space. Benedikt introduces many measures of isovists that gradually change based on position (isovist fields), and demonstrates that these fields are related to how a space is perceived.

Visibility analysis has led to a number of partitioning schemes that all seek to summarize space through segmentation. Hillier and Hanson discuss the concept of \emph{convex maps}, which are a partition of a space into convex regions~\cite{hillier1989social}. Hillier also creates a set of maximal \emph{overlapping} convex shapes which taken together form the entire space~\cite{hillier2007space}. These convex maps are also related to \emph{axial maps}, which determine lines that form a skeleton of a space (reminiscent of the lines that connect regions in BWTA), and induce a graph that represents the space~\cite{turner2005algorithmic}. 

Peponis et al. propose several possible partitionings: \emph{m}-partitions (a partition composed of the minimum number of convex shapes), \emph{s}-partitions (a partition that extends all walls at their interior corners), and \emph{e}-partitions (a partition where a boundary is always crossed whenever a wall becomes visible or disappears from view)~\cite{peponis1997description}.
While this type of analysis could certainly be applied to games, the partitions themselves are overly restrictive: it is useful to allow partition areas to be slightly concave, and there are many discontinuities in the spaces we analyze.

Odawara and Takizawa consider the problem of finding an \emph{isovist cover} of a space. That is, a set of mutually-visible vantage points from which every point in a given space is visible (a slight variation on the Art Gallery problem)~\cite{odawara2025isovist, o1987art}.
They frame this as an optimization problem (minimize the number of viewpoints subject to constraints that every point in the space is visible and the viewpoints are mutually visible). 
This method does create a graph, but the identified segments can overlap and do not necessarily cover the space.

Zhao et al. also identify rooms using agglomerative hierarchical clustering~\cite{zhao2026extraction}. Their distance metric uses a hyperparameter $\lambda$ to trade off the importance of visibility and travel distance. This creates two theoretical issues: when $\lambda$ is small, travel distance dominates and large convex rooms will be needlessly separated. When $\lambda$ is large, this method has the same theoretical problems as other isovist-based methods (see Discussion).

\subsection{Collage Generation}

One use case for our segmentation method is collage-based level generation. By collage-based level generation we refer to a constructive PCG approach for creating level designs by pasting together segments from a pre-existing library. This technique (or variations on it) is fairly common in existing procgen applications. \emph{Spelunky}~\cite{spelunky} and \emph{Dead Cells}~\cite{deadcells} both use a variation on this approach, with hand-authored level chunks that the system can slightly alter before placement~\cite{spelunkygen, deadcellsgen}. \emph{Deep Rock Galactic}~\cite{drg} (a 3D mining / action game) also uses a similar approach, though the chunks can also be transformed and occasionally overlap~\cite{drggen}. Occupancy-Regulated Extension~\cite{occupancy} uses a hand-authored library of smaller movement chunks (together with occupancy annotations) to generate \emph{Super Mario} levels. \emph{Dead Cells} designer S\'{e}bastien Benard summarizes the appeal of this type of generation: ``... they used a hybrid approach between procedural generation and handmade levels, giving them that consistent feeling while maintaining a lot of diversity.''~\cite{deadcellsgen}. Whether or not these collage-based levels really do feel more handcrafted than levels generated with other approaches, the technique remains popular. However, it also requires the author to create these level chunks. In this work, we explore level segmentation as a way to remix an existing level design by automatically identifying appropriate chunks.

\section{Perceptual Segmentation}

\begin{figure*}
    \centering
    \includegraphics[width=0.93\linewidth]{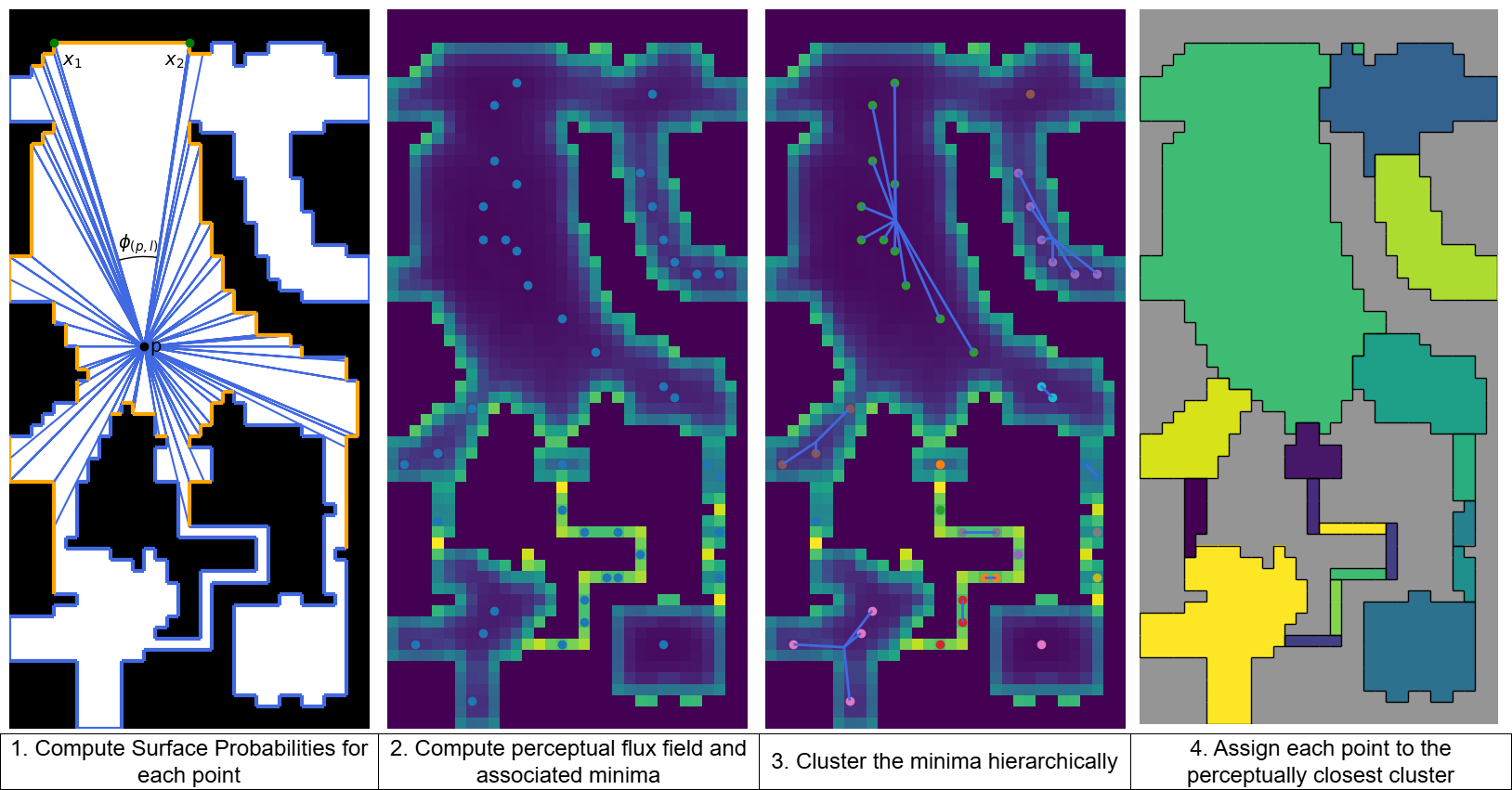}
    \caption{The pipeline for computing the Perceptual Distance Segmentation shown on ``Bubble Mountain'' from \emph{Super Metroid}. We begin by computing the probability of observing (angular area of) each visible surface (orange) from each point. The figure shows an example for a single point. Second, we use the Jensen-Shannon divergence of these probabilities to compute an approximate rate of change at each interior point (perceptual flux). We then find the minima of this field, and cluster the minima hierarchically based on the perceptual distance. Finally, the non-minima points are assigned to the cluster that they are closest to (in terms of perceptual distance). }
    \label{fig:pipeline}
\end{figure*}

So far, we have been vague how to use visibility (straight-line connectedness) information to segment a space.
Our approach is based on agglomerative hierarchical clustering. First, we discretize the space into a grid of points $G$. Initially each point is treated as a separate cluster, and we merge points into clusters based on some measure of the similarity of their visibility information. The intuition behind this is that the points in a convex room are all mutually visible. In a more complex shape, the regions that might be called rooms are approximations of this, and their visibility information should be approximately similar. The main contribution of this paper is an appropriate ``perceptual distance'' measure.

\subsection{Perceptual Distance}

Rather than treating the interior points as the components of the isovist, we note that when observing a space, you actually \emph{see} the places where vision stops: the perimeter of the isovist (Figure~\ref{fig:pipeline}.1 shows an example). We model the act of observation as a random process: a viewer at point $p$ selects a uniform random angle $\theta$. They observe the (unique) point $x = \mathrm{isovist}_p(\theta)$ that is visible in direction $\theta$ from $p$. The perceptual distance between a viewer at $p_1$ and a viewer at $p_2$ is the amount of information that the observed point $x$ gives about the observing point $p$. Treating the observed point $x$ as a probability distribution, we can use a distributional distance to quantify the distance between $p_1$ and $p_2$. In this case, we use the Jensen-Shannon Divergence~\cite{lin1991divergence} because it is  easy to approximate and can be extended to a notion of cluster distance. In order to compute this divergence, we need to calculate the probability of observing each $x$ from both $p_1$ and $p_2$. This section details the process we use to estimate these probabilities.

We define the perceptual distance between two points $D_P(p_1, p_2)$ using the following random variables:
\[ P \sim \mathcal{U}(\{p_1, p_2\})\]
\[ \Theta \sim \mathcal{U}(0, 2\pi)\]
\[ X \sim \mathrm{isovist}_P(\Theta)\]
\[ I \sim 1 - H(P | X)\]

Here $H$ denotes the Shannon entropy. The perceptual distance between $p_1$ and $p_2$ is $E[I]$. To compute $E[I]$, we need to know $Pr(P = p_i | X = x)$ for each observer $p_i$. By Bayes' theorem,
\[ Pr(P = p_1 | X) = \frac{Pr(X | P = p_1)}{Pr(X | P = p_1) + Pr(X | P = p_2)}\]


These probability distributions are not trivial to calculate.\footnote{When $x$ lies on a segment $(a, b)$, $P(X = x | P = p) = \frac{h}{d^2}$ where $h = ||p - \mathrm{proj}_{(a, b)}p||$ and $d = ||x - p||$. So $H(P | X) =  -\sum_{(a, b) \in S}\int_{t = a}^{b} \frac{h}{||p - t||} \log (\frac{h}{||p - t||})$, which is challenging to compute exactly.} Instead, we opt for a discrete approximation: given a set of walls $W$, we discretize these walls into a set of segments $S$. The approximate observation process gives a random \emph{segment} $L$ rather than a random point $X$. For a visible segment $\ell = (x_1, x_2)$, the probability of observing a point on that segment from $p$ is proportional to the angular area taken up by $\ell$ (see Figure~\ref{fig:pipeline}.1). Choices of $\Theta$ that observe $\ell$ would be angles that fall within the angle $\phi_{(p, \ell)}$ formed by the endpoints of the visible part of the segment $(x_1, x_2)$ and the observing point $p$ as the pivot. So:
\[ Pr(L = \ell\ |\ P=p) = \frac{|\phi_{(p, \ell)}|}{2\pi} \]

As the segment size goes to zero (as $x_2 \to x_1$), this quantity is equivalent to the probability of observing $x_1$ from $p$.

Given the segments $S$, we compute this metric using the entropy formula for Jensen-Shannon Divergence. To compute the entropy of a given point, we use

\[ \hat{H}(p) = -\sum_{\ell \in S} Pr(\ell | P=p) \log_2(Pr(\ell| P = p)) \]

This formula generalizes for a set of points $Q = \{ p_1, \ldots p_n\}$. The probability of observing a given segment from a randomly chosen $p_i \in Q$ is:

\[ Pr(s | P \in Q) = \frac{1}{|Q|} \sum_{p_i \in Q} Pr(s | P = p_i)\]

So the entropy of observing from a random point of $Q$ is:

\[ \hat{H}(Q) = - \sum_{\ell \in S} Pr(\ell | P \in Q) \log_2(Pr(\ell | P \in Q))  \]

For two singleton points, we have

\[ \hat{D}_P(p_1, p_2) = \hat{H}(\{p_1, p_2\}) - 0.5 (\hat{H}(p_1) + \hat{H}(p_2)) \]

and for two sets of points,

\[ \hat{D}_P(Q_1, Q_2) = \hat{H}(Q_1 \cup Q_2) - 0.5 (\hat{H}(Q_1) + \hat{H}(Q_2))\]

For a set of points, this distance is equivalent to the previous definition, except we use a random variable $Q \sim \mathcal{U}(\{Q_1, Q_2\})$ to select one of the two clusters, then $P \sim \mathcal{U}(\{q \in Q\})$ selects a uniform point within the cluster. The distance between two clusters is the expected information gained about $Q$ (the choice of cluster) by observing $X = x$ at angle $\Theta$ from $P$. We compute $\hat{D}_P$ exactly (using the segment probabilities) without sampling.

\subsection{Properties of Perceptual Distance}

This definition of perceptual distance has a few useful properties. Firstly, it is relatively simple to compute: we only need to store $\phi_{(p,s)}$ for each grid point $p \in G$ and each segment $s \in S$. We compute these by individually checking the occlusion of each pair of segments. This determines (for each observer point $p$) what the visible segments are and their angular area. More efficient algorithms for computing the isovist (and therefore these angles) exist, but are more complicated to implement~\cite{bungiu2014efficient}. The segment discretization $S$ also makes it simple to compute the probabilities for uniform mixture distributions (which is key for hierarchical clustering). For a mixture, we just need to track the total likelihood for each segment and the total number of points in the mixture. Then, to combine two mixtures, we simply add the segmentwise likelihoods and add the number of points.


The perceptual distance is bounded between 0 bits and 1 bit, and the original definition $D_P$ is scale-invariant. The approximate $\hat{D_P}$ is approximately scale-invariant (dependent on the segment discretization). 
This is useful because hierarchical clustering requires a threshold value at which to stop the merging of clusters. Scale-invariance of the metric allows us to identify a single threshold that will work at multiple scales within a single level.

\subsection{Perceptual Fields and Perceptual Flux}

As with isovist information, our definition of perceptual distance has an associated field that describes the amount of visual change at a given point. 
For a given point $p$, we can estimate the total perceptual change at $p$ by

\[ |\nabla D| \approx \frac{1}{|C|}\sum_{d \in C} \frac{\hat{D}(p, p+d)}{||d||} \]

In this case, we limit $C$ to the cardinal directions for convenience: this formula is then the average perceptual distance of $p$ to its neighbors in the grid $G$. We term this value the \emph{perceptual flux} at $p$ since it estimates the total perceptual change that happens at $p$. Figure~\ref{fig:pipeline}.2 shows this flux field for ``Bubble Mountain'', a room from \emph{Super Metroid}.

In general, the perceptual flux will be high near walls since each point near a wall has a highly-concentrated probability of observing a point on the nearby wall. It is also high at the open boundary of a room since unseen walls become visible once the threshold of the room is crossed. In this way, local minima of the flux field correspond to room centers. Figure~\ref{fig:pipeline}.2 shows an example of these minima.

\subsection{Level Segmentation with Perceptual Distance}

In order to use this perceptual distance metric to segment a space, we start by discretizing the space and computing the set of visible segments for each point of the space. We then compute the perceptual flux field (as defined above), and compute the local minima of that field. We use an agglomerative hierarchical clustering algorithm to segment these minima~\cite{ahc2012}. The distance metric for this clustering is the perceptual distance, and the threshold is 0.59. 
Each non-minima point is then assigned to the cluster that it is closest to in terms of perceptual distance. This algorithm, which we call \emph{Perceptual Distance Segmentation} (PDS), groups the points of a space by the similarity of what they can see.

\section{Perceptual Segmentation Experiments}

In this section we apply the segmentation algorithm and show the resulting segmentations. We first apply it to author-designed rooms in order to determine an appropriate threshold value. We then apply it to \emph{StarCraft} maps in order to compare the clusterings with prior work. Finally, we apply it to rooms from \emph{Super Metroid}.

\subsection{Threshold Selection}

\begin{figure}
    \centering
    \includegraphics[width=0.69\linewidth]{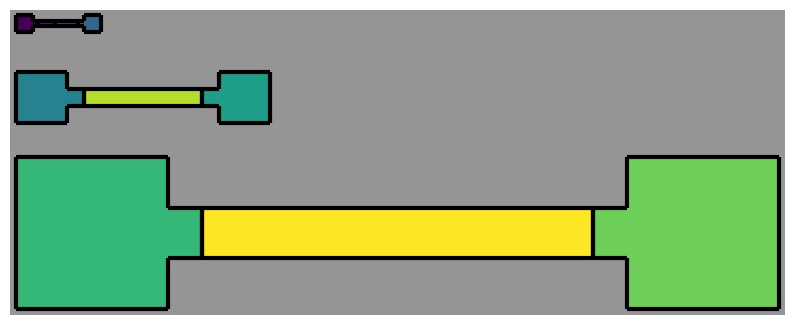}
    \includegraphics[width=0.29\linewidth]{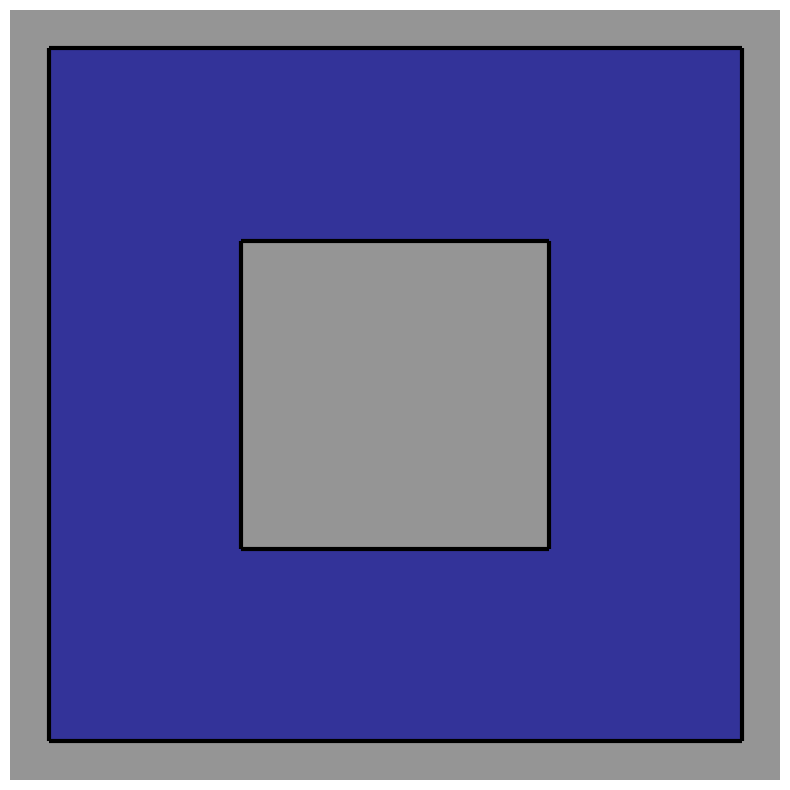}
    \caption{Left: Three scaled versions of the same figure, clustered with perceptual distance. The threshold of 0.56 bits is the lowest threshold that does not separate either hallway at the two larger scales. Right: a room with an interior obstacle which is clustered with perceptual distance into a single cluster. The threshold of 0.59 bits is the lowest threshold that does not separate this room into multiple clusters.}
    \label{fig:thresholds}
\end{figure}

In order to select an appropriate threshold, we designed two segmentation scenarios, decided on a canonical segmentation for each, and tuned the threshold to match this segmentation. In the Hallway example (Figure~\ref{fig:thresholds}), we decided that the segmentation should segment the two rooms and the hallway separately at multiple scales. The lowest threshold that produces this segmentation is 0.56 bits. In the Ring example (Figure~\ref{fig:thresholds}), we decided that the segmentation should produce a single cluster (despite the obstacle in the center of the room). The lowest threshold that produces our desired segmentation is 0.59 bits. These examples are both arbitrary; as with most unsupervised clustering methods, the choice of a threshold or number of clusters is more of an art than a science. 
We have found that numbers between 0.5 bits and 0.6 bits tend to work well in practice across multiple domains. Since the perceptual distance is approximately scale-invariant, we use the same thresholding value (0.59 bits) across all of our evaluations.

\subsection{Comparison to Prior Work on \emph{Starcraft}}

\begin{figure}
    \centering
    \begin{subfigure} {0.48\linewidth}\includegraphics[width=\linewidth]{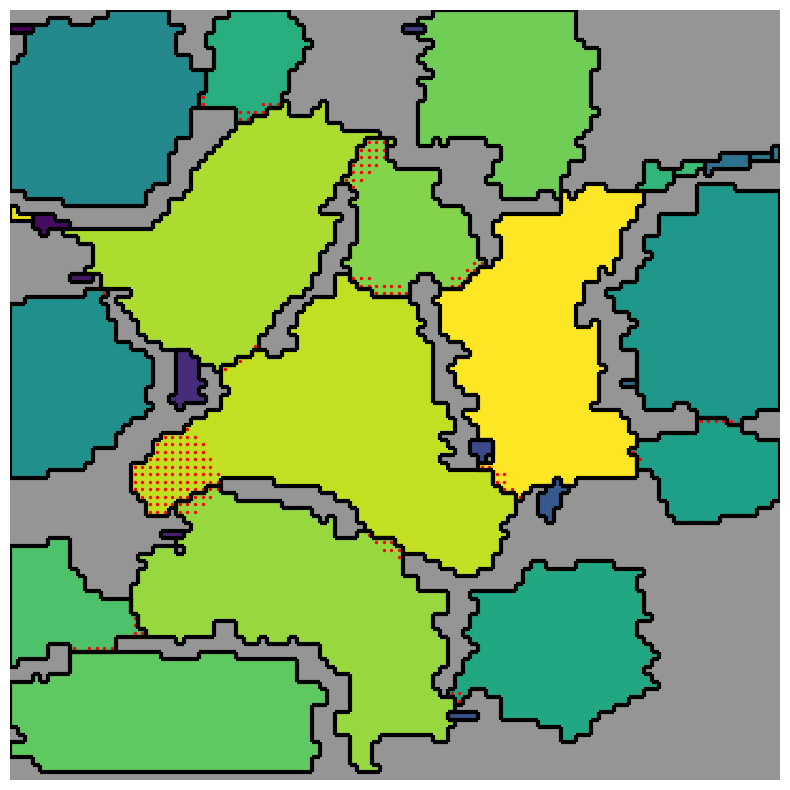}
    \caption{PDS}
    \end{subfigure}
    \begin{subfigure}{0.48\linewidth}
    \includegraphics[width=\linewidth]{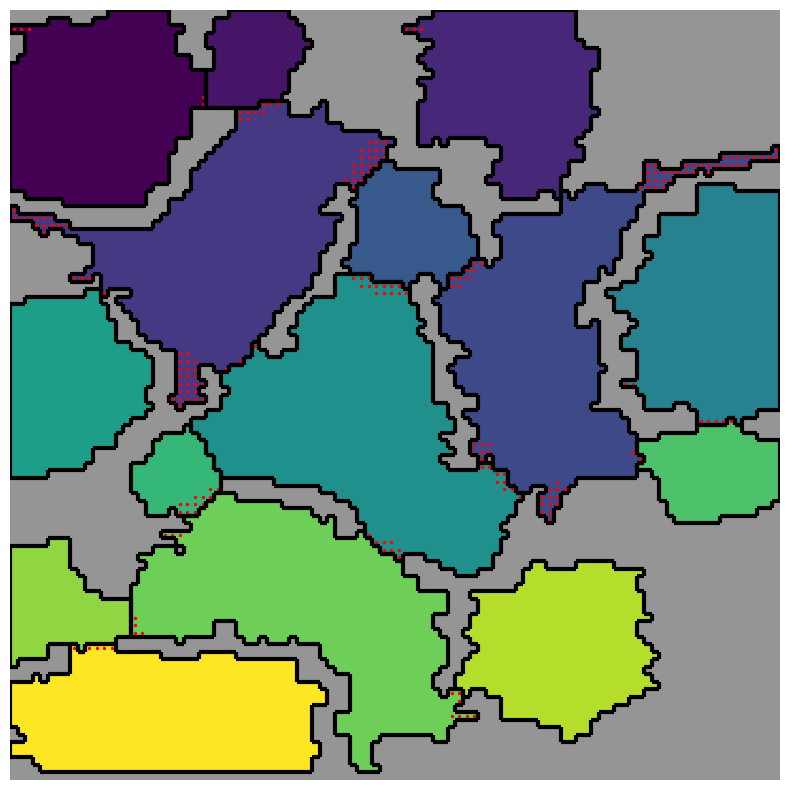}
    \caption{BWTA}
    \end{subfigure}
    \caption{Clusterings for the \emph{StarCraft} map Longinus. Longinus is the map from our sample with the highest cluster agreement score (0.96). Red shaded regions indicate cluster disagreement. The primary difference is that BWTA creates an additional cluster near the center. BWTA does not split the small top-right spur into multiple segments}
    \label{fig:longinus}
\end{figure}

Perkins' work on BWTA for \emph{Starcraft} is the most closely-related prior work and their original output data is available.\footnote{\url{https://code.google.com/archive/p/bwta/}} In this subsection we compare directly to their results to characterize our approach~\cite{perkins2010terrain}. Perkins validated their system with human expert data which is not publicly available: we instead compare segmentations produced via perceptual distance directly to segmentations produced by BWTA. Although the perceptual distance is fundamentally polygon-based, we do not have access to the polygonalization of the StarCraft maps used by BWTA. Instead, we use a discretized (treating each pixel as a square made up of four surfaces) downsampled ($0.1$ scale) version of the maps, comparing against a discretized and downsampled version of the BWTA output image. This degree of downsampling is justified because the BWTA output images are larger than the underlying map (e.g. \emph{Andromeda} output is $1024 \times 1024$ pixels, while the underlying map is only $128 \times 128$ tiles). We compare against the first 24 of the 48 BWTA maps. This is an imperfect point of comparison, but it serves to characterize our output segmentations.

\begin{figure}
    \centering
    \begin{subfigure}{0.48\linewidth}
    \includegraphics[width=\linewidth]{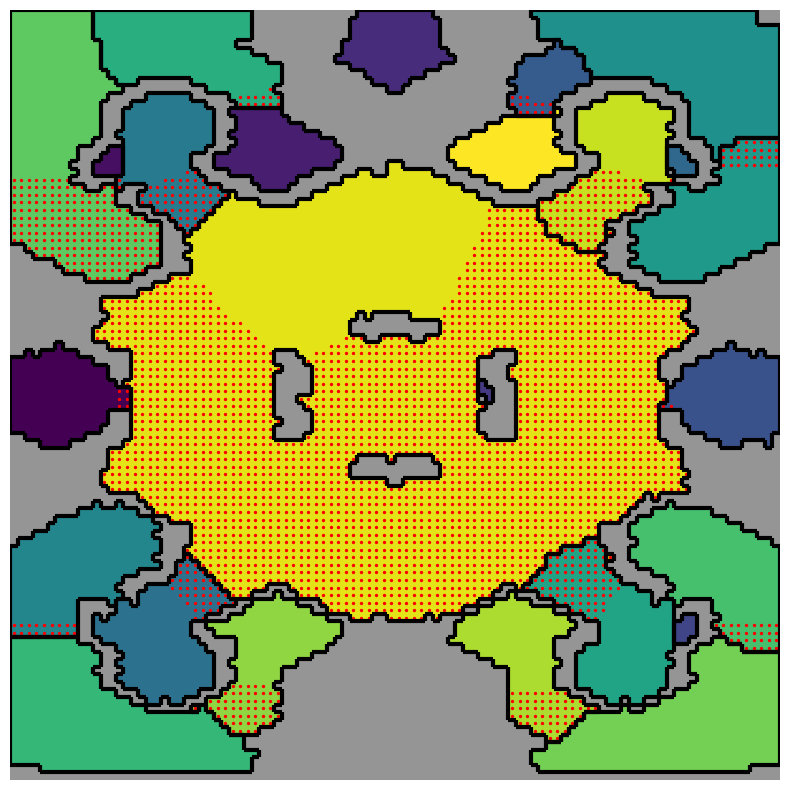}
    \caption{PDS}
    \end{subfigure}
    \begin{subfigure}{0.48\linewidth}
    \includegraphics[width=\linewidth]{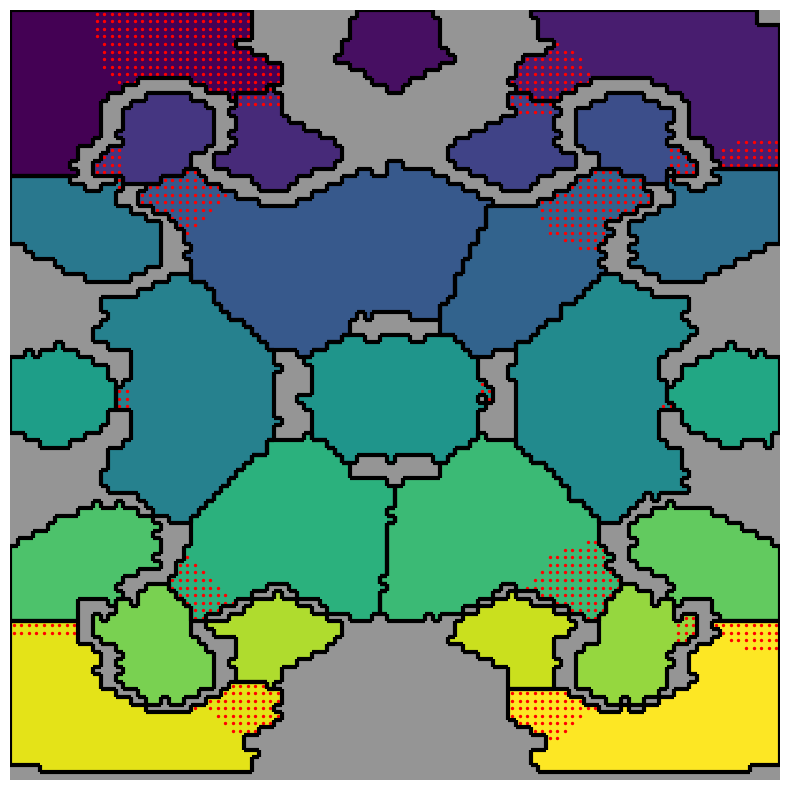}
    \caption{BWTA}
    \end{subfigure}
    \caption{Clusterings for the \emph{StarCraft} map Andromeda~1.0. Andromeda~1.0 is the map from our sample with the lowest cluster agreement score (0.71). Red shaded regions indicate cluster disagreement. PDS creates a large central cluster, whereas BWTA breaks this central cluster into smaller parts.}
    \label{fig:andromeda}
\end{figure}

Figures~\ref{fig:longinus} and \ref{fig:andromeda} show comparisons between the segmentation produced by PDS (ours) and the original BWTA method. There are many existing methods for comparing the quality of clusterings, but most of them make inappropriate assumptions for this domain (for example, silhouette score with Euclidean distance can reward the assignment of a point to a cluster center that is behind a wall). In order to compare these clusterings, we define a cluster agreement score. For clusterings $\mathbb{C} = \{C_i\}$ such that the $C_i$ form a partition of a space $S$:

\[ A(\mathbb{C}_1, \mathbb{C}_2) = 1 - \left( \frac{1}{|S|} \sum_{C_1 \in \mathbb{C}_1} |C_1 \setminus \underset{{C_2 \in \mathbb{C}_2}}{\mathrm{argmax}} (C_1 \cap C_2) | \right) \]

Where $|S|$ is the total size of the clustered space. This definition assigns each cluster of $C_1$ to the cluster of $C_2$ that has the most overlap, then counts the total fraction of points that are not in this overlap. This score is asymmetric: it will be low if $\mathbb{C}_1$ is close to a sub-partition of $\mathbb{C}_2$ (if $\mathbb{C}_1$ is provinces and $\mathbb{C}_2$ is countries, for example). The two-sided score $S(\mathbb{C}_1, \mathbb{C}_2) = \frac{1}{2}( A(\mathbb{C}_1, \mathbb{C}_2) + A(\mathbb{C}_2, \mathbb{C}_1))$ is symmetric. If $C_1 = C_2$, then $A(C_1, C_2) = 1$.

\begin{table}[]
    \centering
    \begin{tabular}{|c|c|}
        \hline
        Metric & Value \\
        \hline
        $A(\mathrm{PDS}, \mathrm{BWTA})$ & $0.83 \pm 0.10$ \\
        $A(\mathrm{BWTA}, \mathrm{PDS})$ & $0.88 \pm 0.03$ \\
        $S(\mathrm{PDS}, \mathrm{BWTA})$ & $0.86 \pm 0.05$ \\
        \hline
        $S(\mathrm{Random}, \mathrm{BWTA})$ & $0.17 \pm 0.06$ \\
        \hline
    \end{tabular}
    \caption{Average cluster agreement values with standard deviation comparing PDS with BTWA segmentations. The Random entry compares BWTA to random segmentations to obtain a baseline. Overall the segmentation agreement scores are high ($86\%$ of points agree with their assigned cluster).} 
    \label{tab:numbers}
\end{table}

Table~\ref{tab:numbers} shows the average cluster agreement between PDS and BWTA. It is useful to distinguish between the two types of agreement. As shown in Figure~\ref{fig:longinus}, the additional cluster created by BWTA in the center contributes 100 tiles to $A(\mathrm{PDS}, \mathrm{BWTA})$ but only 12 tiles to $A(\mathrm{BWTA}, \mathrm{PDS})$ because it can be explained as part of the larger cluster created by PDS. The high average agreement shows that both methods largely agree on most boundaries, though there are cases where they interpret the space extremely differently (as shown in Figure~\ref{fig:andromeda}).


\subsection{Results on \emph{Super Metroid}}

\begin{figure}
    \centering
    \resizebox{0.9\linewidth}{!}{
    \includegraphics[height=3cm]{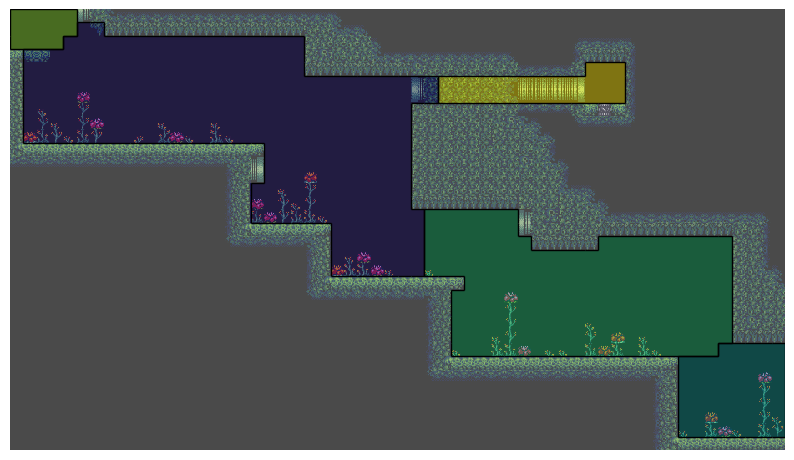}
    \includegraphics[height=3cm]{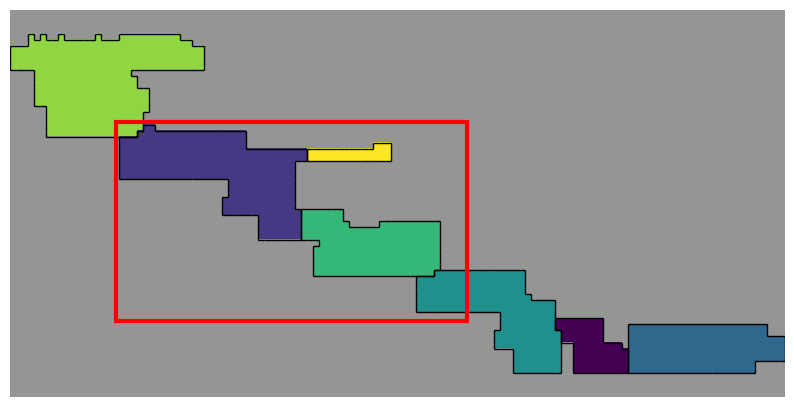}
    }
    \caption{Detail from the generated clustering for the ``Green Hill Zone'' (room \texttt{0x79E52}).}
    \label{fig:ghz_partition}
\end{figure}

\begin{figure}
    \centering
    \resizebox{0.9\linewidth}{!}{
    \includegraphics[height=3cm]{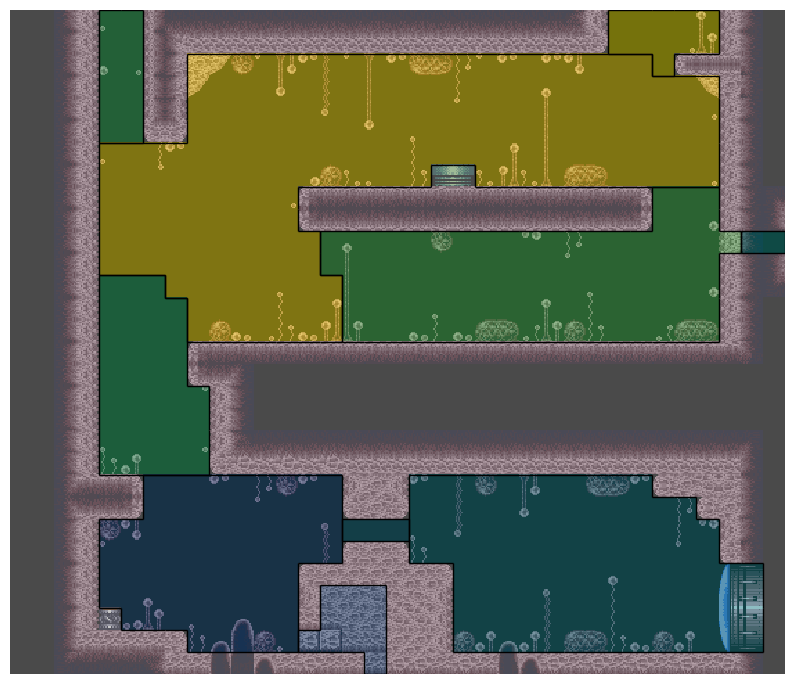}
    \includegraphics[height=3cm]{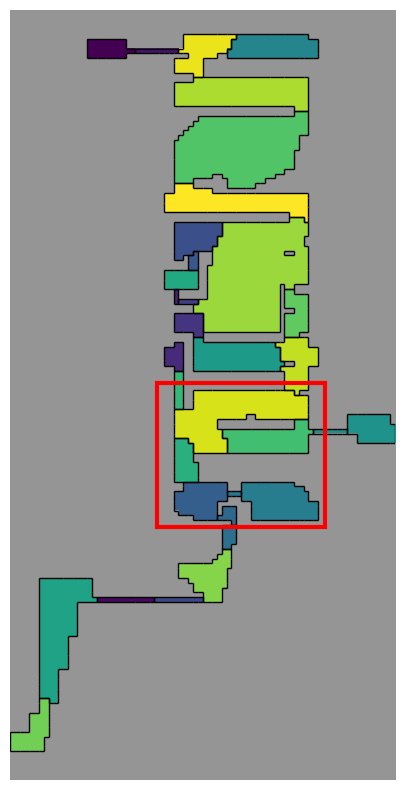}
    }
    \caption{Detail from the generated clustering for ``Big Pink'' (room \texttt{0x79D19}).}
    \label{fig:bp_partition}
\end{figure}

\begin{figure}
    \centering
    \resizebox{0.9\linewidth}{!}{
    \includegraphics[height=3cm]{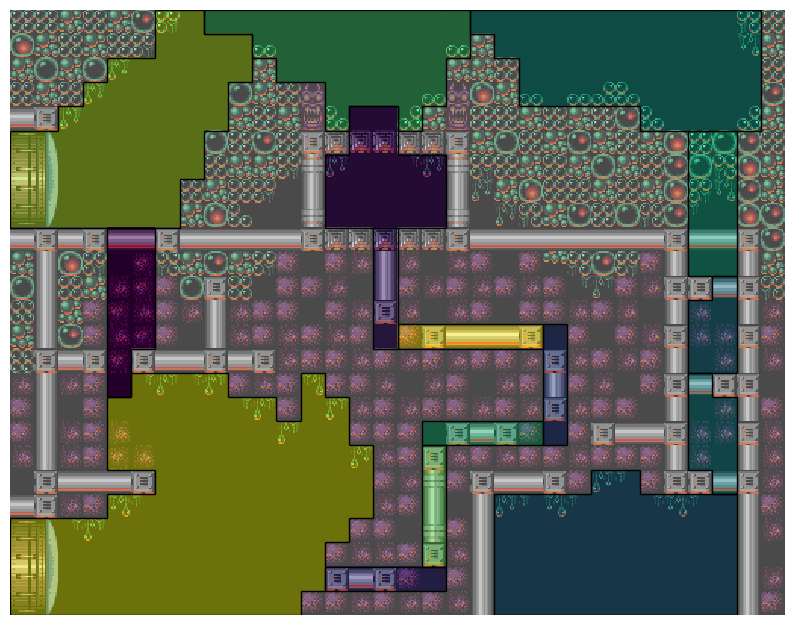}
    \includegraphics[height=3cm]{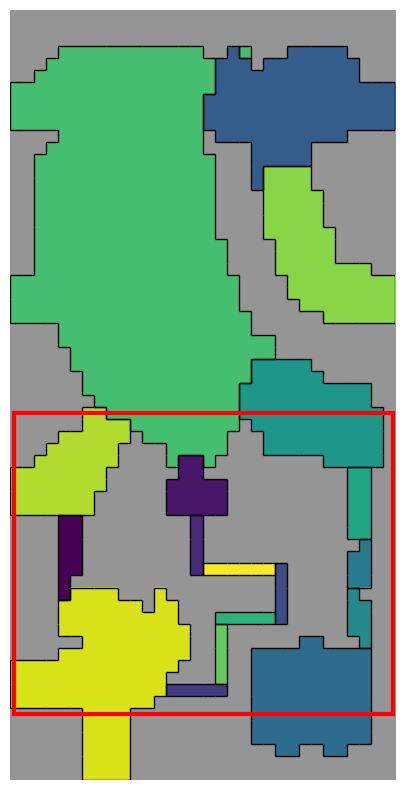}
    }
    \caption{Detail from the generated clustering for ``Bubble Mountain'' (room \texttt{0x7ACB3}).}
    \label{fig:bm_partition}
\end{figure}

\emph{Super Metroid} is an eponymous \emph{metroidvania} (or \emph{search-action}) game: a 2D platformer composed of 255 individual rooms. While some of these rooms are small and simple, many are larger and made up of smaller identifiable areas. We use a parser\footnote{\url{https://github.com/aremath/sm_rando/tree/master/rom_tools}} to find rooms and decompress associated level data. We refer to each room both by the colloquial name and its memory address. We use the Layer 1 (physical geometry) tile type to distinguish between solid tiles and empty tiles. We treat solid, treadmill, spike, and grapple tiles as solid, and all other tiles as empty. This has the limitation that slopes and other tiles that are partially solid are treated as empty, but it has the advantage that destructible tiles that the player can move through are also treated as empty. We selected individual rooms that have enough internal structure to show the potential utility of segmentation.

Figures~\ref{fig:ghz_partition}, \ref{fig:bp_partition}, and~\ref{fig:bm_partition} show example clusterings computed for individual rooms from \emph{Super Metroid} (also using a threshold value of 0.59 bits). As with \emph{StarCraft}, the rooms exhibit complex structures with different scales. These clusterings demonstrate some features of Perceptual Distance Segmentation. The ``Green Hill Zone'' segmentation (Figure~\ref{fig:ghz_partition}) shows how PDS splits off smaller side passages. While this segmentation is only one interpretation of the space, it does break the other rooms apart at geometric bottlenecks in the room. The ``Big Pink'' segmentation (Figure~\ref{fig:bp_partition}) shows how PDS splits apart an ambiguous space. In this case, allowing concave rooms creates a more parsimonious breakdown. The ``Bubble Mountain'' partition (Figure~\ref{fig:bm_partition}) shows how smaller-scale structures can be detected at the same time as larger ones. All three passages between the upper and lower parts of this room are broken up into their smaller components, while the main portion of the upper part of this room remains intact.

There are many different ways to segment these spaces, and the PDS segmentation will not always align with human intuition. While the PDS segmentations are only one interpretation of the underlying space, these results show that this single approach can produce reasonable interpretations with a single choice of hyperparameter.

\subsection{Ablations}

We also tried two variations of PDS. First, we used hierarchical clustering on the entire set of points, replacing steps 3-4 from Figure~\ref{fig:pipeline} with one big hierarchical clustering step (referred to as No Minima or NM). Second, we tried improving the perceptual distance estimate by using a surface discretization which is half as large (Smaller Surfaces or SS). Initial results on ``Bubble Mountain'' suggest that neither of these changes makes a big difference in the output segmentations. No Minima is more than $10\times$ slower than standard PDS, and $S(\mathrm{PDS}, \mathrm{NM}) = 0.93$. Likewise $S(\mathrm{PDS}, \mathrm{SS}) = 0.96$. While this result only covers one example, it helps to justify both of these algorithm design choices.

\section{Collage Generation using Perceptual Distance}

One use case for these clusterings is to use them as level segments for procedural level design. We demonstrate an example of this using the segments obtained from \emph{Super Metroid}.

The collage generation algorithm is simple. To start, PDS breaks up each level design into segments. We then start from a random seed segment, and repeatedly iterate through the remaining segments until we find a segment that matches one of the seed's borders. We attach that segment in the selected location, and recompute the borders for the in-progress level design. This recomputation step is significant since a single placement can create a new border shape which can then be matched with a different segment on a later step. This process continues until either a) the set of segments is exhausted, b) the desired number of segments have been added, or c) none of the remaining segments match any of the existing borders.

The main challenge here is determining what counts as a valid border match. We treat each identified segment as also containing the solid tiles that immediately surround it (with overlap allowed). A segment's \emph{borders} are the contiguous regions where the internal tiles of the segment are adjacent to (non-solid) tiles of a different segment. For each border, we identify an \emph{interior} and an \emph{exterior} face, and the direction(s) that the border faces. Only borders that face in opposite directions are compatible.

\begin{figure}
    \centering
    \includegraphics[width=0.8\linewidth]{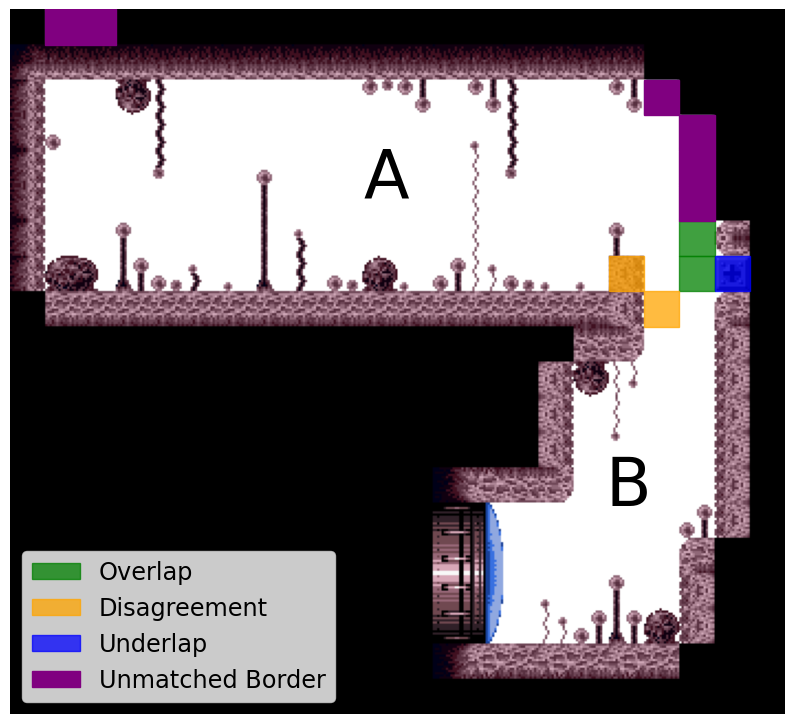}
    \caption{Example collage border match. \emph{Overlap} (good) refers to locations where the inner and outer borders of the two rooms overlap. \emph{Disagreement} (bad) refers to locations where one room has a solid tile and the other has an air tile (or vice versa). \emph{Underlap} (bad) is the set of locations of either border that are not covered by the other room. \emph{Border} tiles are the remaining additional borders of the current collage. The border in the top left is due to hidden tiles: the ceiling there looks solid but does not have collision.}
    \label{fig:collage_placement}
\end{figure}

When attaching a segment $S_1$ to a segment $S_2$, we attach them at one of their borders $B_1$ and $B_2$. This border matching is an optimization problem, illustrated in Figure~\ref{fig:collage_placement}.

We wish to maximize the border \emph{overlap}: the interior face of $B_1$ should overlap with the exterior face of $B_2$ and vice versa. 
However, there are three things that can go wrong with a placement. If $S_1$ and $S_2$ \emph{disagree} about the placement of solid and air tiles, this creates a possible mismatch. If the placement of $S_2$ leaves empty space in the exterior border of $B_1$, the two rooms \emph{underlap}. This is a problem because it leaves an open border region which might be impossible to match with an existing room. Finally, if the placement of $S_2$ \emph{conflicts} with an already-placed room, there might be disagreement between the tiles each room has on those locations. If this overlap is large, then a new unintended entrance could be opened. The \emph{badness} of a placement is the total number of these violations.

\begin{figure*}
    \centering
    \includegraphics[width=0.7\linewidth]{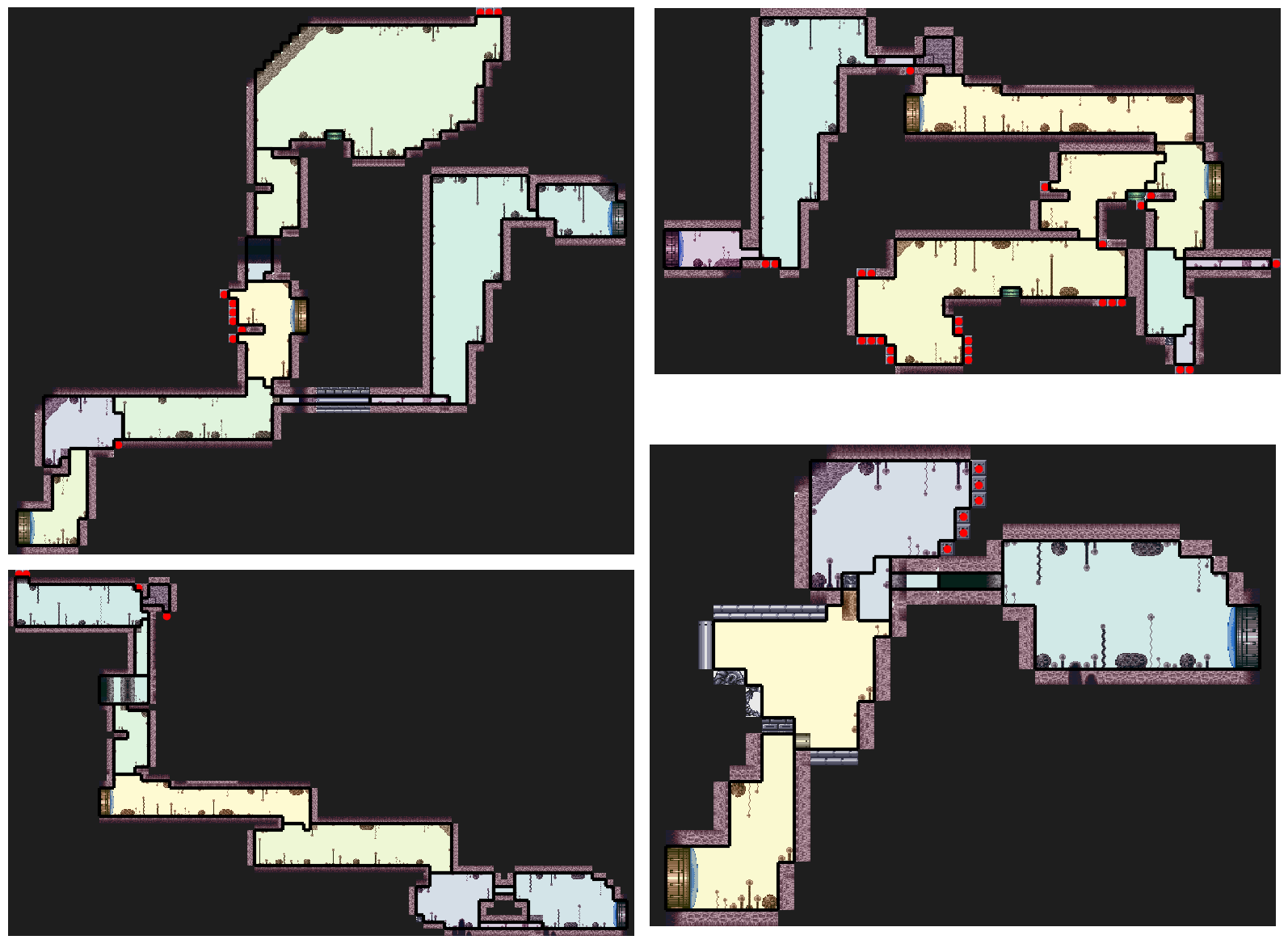}
    \caption{Randomly-selected generated collage outputs from ``Big Pink'' (seeds 343, 344, 345, and 346). Sub-rooms are color-coded based on the segmentation from Figure~\ref{fig:bp_partition}. Red dots indicate remaining unfilled borders. Each generated output uses a badness threshold of 3 and adds up to 10 rooms. In the bottom right (seed 346), an awkward placement ends the generation at only 6 rooms: there are no valid matches for the remaining border.}
    \label{fig:collages}
\end{figure*}

When considering a placement, we constrain the badness to be below a threshold $B$. Allowing more violations creates a more permissive set of placements, which can be both good or bad: more allowed placements increases the possibility space of the generator, but some of these placements may be awkward or leave empty space that cannot be filled by an existing segment.

\subsection{Collage Generation Results}

Figure~\ref{fig:collages} shows four example generated level designs using the segmentation for ``Big Pink'' (room \texttt{0x79D19}). This room is comprised of 34 segments, with 73 total borders. The average (interior) border size is $2.4\pm1.5$ tiles. The generated examples shown in Figure~\ref{fig:collages} use a badness threshold of $B=3$. At this threshold, $21\%$ of all border pairs are compatible (out of a possible $40\%$ when $B=18$).\footnote{Diminishing returns in compatibility start at $B=5$. However, some of the $B=4$ placements simply look wrong, so we selected $B=3$.}


The generated outputs show that the algorithm is capable of creating a wide variety of possible room shapes with different amounts of internal branching. Since sub-rooms are created by simply pasting each segment, the outputs have many places where neighboring tiles do not match graphically. We do not re-tile these rooms, but algorithms like WFC would be a good fit for this task~\cite{wfc}.

Although we do not check gameplay properties of the outputs, these examples all look navigable. There are places where they are graphically disconnected, but these are all either destructible blocks or tiles that do not have collision. These examples show how remixing existing content can still create a wide variety of different designs.

\begin{figure}
    \centering
    \includegraphics[width=0.7\linewidth]{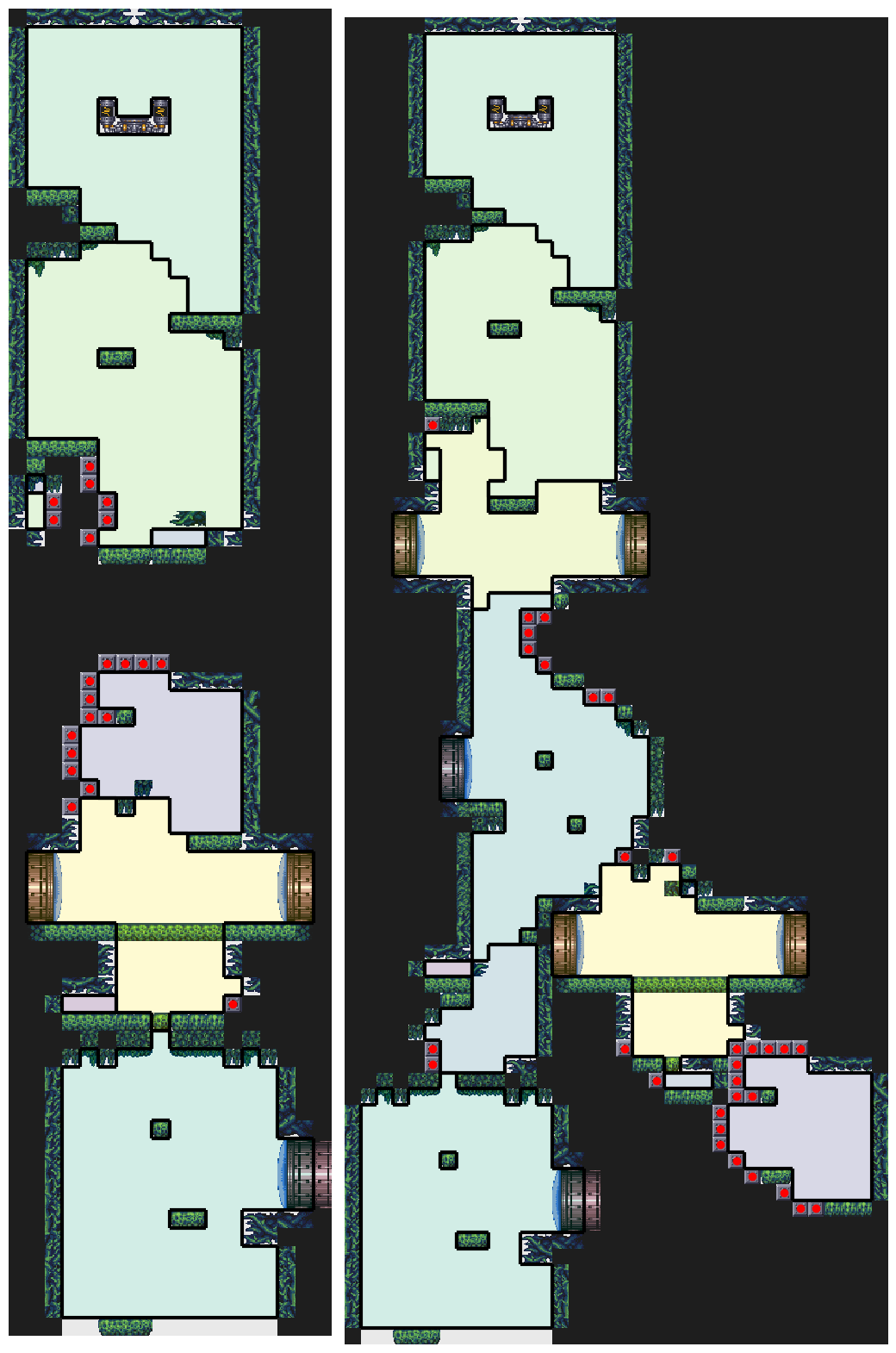}
    \caption{Randomly-selected generated collage outputs from ``Green Brinstar Entrance'' (room \texttt{0x79AD9}) (seeds 343 and 344). Seed 343 is disconnected because one of the identified clusters was disconnected.}
    \label{fig:gbe_collage}
\end{figure}


\section{Discussion}

In this section we discuss alternatives to PDS, its limitations, and possible future work.

\subsection{Alternative Approaches}

\begin{figure}[t]
    \centering
    \begin{subfigure}{0.49\linewidth}
    \includegraphics[width=\linewidth]{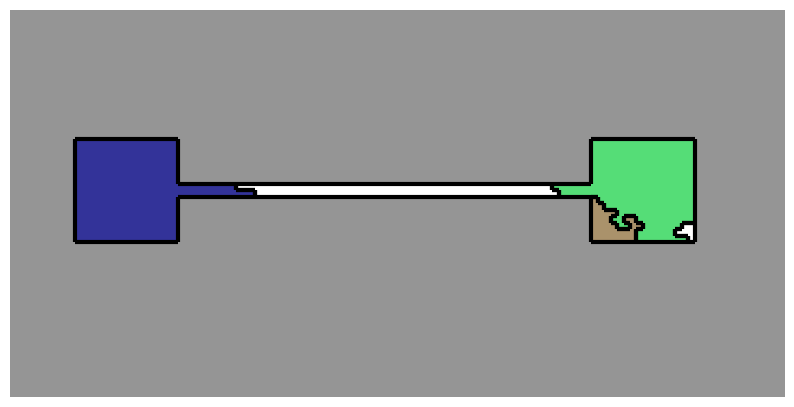}
    \caption{Isovist Jaccard Index (IJI)}
    \end{subfigure}
    \begin{subfigure}{0.49\linewidth}
    \includegraphics[width=\linewidth]{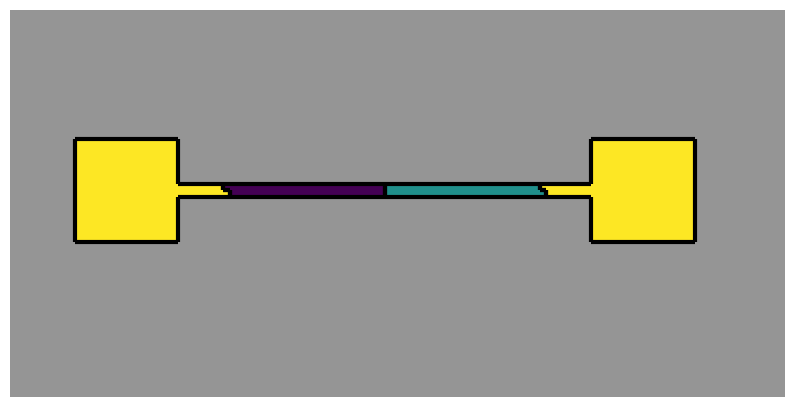}
    \caption{PDS}
    \end{subfigure}
    \caption{A long hallway segmented with two different methods. In IJI, clusters below size 6 are also merged and the threshold value of 0.4 is chosen manually by the authors for best results. For PDS, the threshold value of 0.59 is the same as our other domains. While the PD segmentation splits the hallway in two and does have some intrusion of the rooms into the hallway, the IJI segmentation also has some segments within a room due to the sudden increase in the area of the isovist once the hallway is visible.}
    \label{fig:long_hallway}
\end{figure}

\begin{figure}
    \centering
    \includegraphics[width=\linewidth]{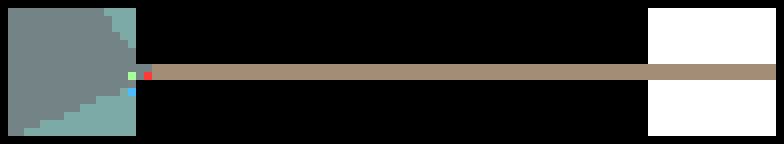}
    \caption{The Long Hallway Problem. Moving from the red point (inside the hallway) to the green point (inside the room) allows a viewer to see the two interior corners of the room which were previously invisible. Moving from the blue point (inside the room) to the green point allows the viewer to see the hallway. If the hallway is sufficiently long, interior visibility approaches will treat the blue point as more dissimilar from the green point than the red point even though the green and blue points are both in the same room.}
    \label{fig:long_hallway2}
\end{figure}

In addition to PDS, we implemented various approaches based on the interior points of the isovist. For example, in one approach, the distance metric we used was the normalized symmetric set difference (Jaccard index~\cite{jaccard1912distribution}) between theisovists. This metric treats the isovists of each observer point as a set of points, and computes the union between these sets. When the union is large compared to the whole set of visible points, the points are perceptually close to each other. Figure~\ref{fig:long_hallway} illustrates a fundamental problem with this method: the ``Long Hallway Problem'' (explained further in Figure~\ref{fig:long_hallway2}). Other methods based on interior visibility (e.g. Zhao et al.'s) have the same issue~\cite{zhao2026extraction}. PDS avoids this problem by considering the visibility of surfaces rather than interior points, and by using a metric scales surface importance based on distance.

We also implemented an approach using the visibility graph directly. In this setting, the points of the space are treated as a graph where the unweighted edges indicate mutual visibility. We used the Leiden community-detection algorithm~\cite{traag2019louvain} on this graph to create a segmentation. This algorithm can correctly characterize this example as two rooms connected by a hallway, but the segmentation depends on a \emph{resolution} hyperparameter. In experiments on \emph{Super Metroid}, different settings of resolution are effective on different parts of each room, making this method difficult to use on large and varied rooms.

\subsection{Limitations}

Perceptual Distance Segmentation has several limitations. Firstly, it is worth questioning the wisdom of using visibility-based segmentation in 2D games: the graphics of these games and their gameplay do not depend at all on the notion of visibility that we use to compute the segmentations. With full knowledge of a game's dynamics it might be possible to develop a more appropriate segmentation based on bottlenecks in possible player trajectories. Secondly, the current implementation of our approach is quite slow. In a map with $n$ tiles, there are typically $O(n)$ segments. PDS runs an $O(n^2)$ algorithm to determine the visible part of each segment for each point, so the asymptotic runtime is $O(s^6)$ where $s$ is the side length of a square level. While there are multiple ways that this could be improved, the average running time for \emph{StarCraft} segmentation was $2.4 \pm 0.8$ hours per map, or $1.19 \pm 0.4$ seconds per empty tile in the $102 \times 102$ maps. Almost all ($84\%$) of this time is spent computing the surface probabilities. The \emph{Super Metroid} examples are smaller and have more confined spaces: they all finished in under 4 minutes. Without algorithm improvements, it would be difficult to apply this method at larger scales (or in 3 dimensions).

Finally, the collage algorithm presented here is simply a proof of concept: we do not check any gameplay properties of the output levels. The generation process also does not encourage the creation of cycles among placed segments even though the original design usually contains cycles. Cycles can be created (e.g. at the bottom of the bottom-left room in Figure~\ref{fig:collages}) but this is more by accident than design. The generated levels are usually incomplete (some border regions will not be filled), requiring a strategy for filling these spaces. Finally, there are some rooms where the collage outputs are not satisfactory (see Figure~\ref{fig:gbe_collage}). In this case, the segmentation identified some disconnected segments and some borders only have one valid match.

\subsection{Future Work}

The notion of perceptual distance defined here can be extended to 3D space. Surfaces can be discretized into triangles, and the perceptual distance could be computed by treating the angular area of each triangle as a probability distribution. The same segmentation algorithm could be used to analyze 3D structures.

Another issue with the current algorithm is analyzing concave spaces that curve smoothly without a well-defined edge. For example, an S-shaped room would currently be split into multiple segments. 
A clustering algorithm based on perceptual flux at the edge between two clusters might be able to create a single cluster for such a room.

Finally, the perceptual flux is a useful way to characterize space. A level generation algorithm could explicitly try to match a specific distribution of fluxes, emulating the building design technique envisioned by Benedikt in 1979!

\section{Conclusion}

In this paper we have used architectural analysis techniques to create Perceptual Distance Segmentation, a segmentation algorithm for 2D maps. We have applied it to level designs from \emph{StarCraft} and \emph{Super Metroid} to obtain segmentations that summarize the high-level structure of these spaces. This analysis technique forms the basis for a proof-of-concept collage level generator that can generate remixes of existing levels, using only a single initial level design as input. Patterns in level designs exist at many scales. Understanding these patterns is an important way to understand the design process, and this method is one part of that overarching goal.

\section*{Acknowledgements}

This work was funded by the Canada CIFAR AI Chairs Program. We acknowledge the support of the Alberta Machine Intelligence Institute (Amii).

\bibliography{aaai2026}

\end{document}